\documentclass[journal]{IEEEtran}

\ifCLASSINFOpdf
  \usepackage[pdftex]{graphicx}
\else
  \usepackage[dvips]{graphicx}
\fi
\usepackage{amsmath}
\usepackage{amsfonts}
\usepackage{amssymb}
\usepackage{amsthm}
\usepackage{algorithmic}
\usepackage{algorithm}
\usepackage[export]{adjustbox}
\usepackage{booktabs}

\usepackage{array}
\usepackage{tabularx}
\usepackage{multirow}
\usepackage{rotating}
\usepackage{array}

\usepackage[caption=false,font=footnotesize]{subfig}

\usepackage{changepage}

\usepackage{url}
\usepackage{siunitx}
\begin{document}

\title{Integer Optimization of CT-Trajectories using a Discrete Data Completeness Formulation}
\pagenumbering{gobble}

\author{Linda-Sophie~Schneider\thanks{L. Schneider and A. Maier are with the Pattern Recognition Lab, Friedrich-
		Alexander University Erlangen-Nuremberg, Erlangen, Germany (e-mail:
		linda-sophie.schneider@fau.de)},~Gabriel~Herl\thanks{G. Herl is with the Deggendorf Institute of Technology},
        Andreas~Maier,~\IEEEmembership{Member,~IEEE} }

\maketitle

\begin{abstract}
X-ray computed tomography (CT) plays a key role in digitizing three-dimensional structures for a wide range of medical and industrial applications. Traditional CT systems often rely on standard circular and helical scan trajectories, which may not be optimal for challenging scenarios involving large objects, complex structures, or resource constraints. In response to these challenges, we are exploring the potential of twin robotic CT systems, which offer the flexibility to acquire projections from arbitrary views around the object of interest. Ensuring complete and mathematically sound reconstructions becomes critical in such systems. 
In this work, we present an integer programming-based CT trajectory optimization method. Utilizing discrete data completeness conditions, we formulate an optimization problem to select an optimized set of projections. This approach enforces data completeness and considers absorption-based metrics for reliability evaluation. We compare our method with an equidistant circular CT trajectory and a greedy approach. While greedy already performs well in some cases, we provide a way to improve greedy-based projection selection using an integer optimization approach. Our approach improves CT trajectories and quantifies the optimality of the solution in terms of an optimality gap.  
\end{abstract}


\IEEEpeerreviewmaketitle

\section{Introduction}
\IEEEPARstart{X}{-ray} computed tomography (CT) is an important tool in both medical and industrial domains for digitizing three-dimensional structures. It facilitates critical applications such as medical diagnostics and industrial quality control. Traditional industrial CT systems often use circular and helical scan trajectories due to their simplicity and compatibility with fast reconstruction algorithms. However, these standard trajectories may not be ideal for large-scale objects, complex structures, or resource constraints \cite{herl2020scanning,bauer2020scan}. Furthermore, mechanical instabilities further complicate the acquisition \cite{maier2011analysis}. Combining standard circular or helical trajectories with cone beam setups can lead to artifacts due to violations of the Tuy-Smith condition \cite{smith1985image}. These degrade CT reconstruction quality, especially for complex or highly attenuating materials. This highlights the need for more adaptive and optimized CT trajectory planning approaches that consider the unique characteristics of the object being scanned.

Recent advances in CT technology have introduced twin robotic CT systems, where individual robots control the X-ray source and detector. This innovation provides flexibility by allowing projections from arbitrary views around the object of interest \cite{landstorfer2019investigation}. Robot-supported CT systems offer the potential for more tailored and complex CT trajectories, enabling efficient scanning of large objects \cite{holub2019roboct}, metal artifact reduction \cite{herl2020scanning}, and shorter scan times \cite{bauer2020scan}.

The flexibility of robotic CT systems presents challenges. Ensuring mathematically complete reconstructions becomes crucial, requiring sufficient information from the acquired projections. In the standard CT process, users select a continuous curve, often circular or helical, around the region of interest and sample it into equidistant views based on the desired number of projections. Focusing on specific regions of objects typically requires manual CT trajectory planning. While effective, this manual approach is time-consuming, subjective, and may result in suboptimal image quality.
\begin{figure}[!t]
\centering
    {\includegraphics[width=0.55\linewidth]{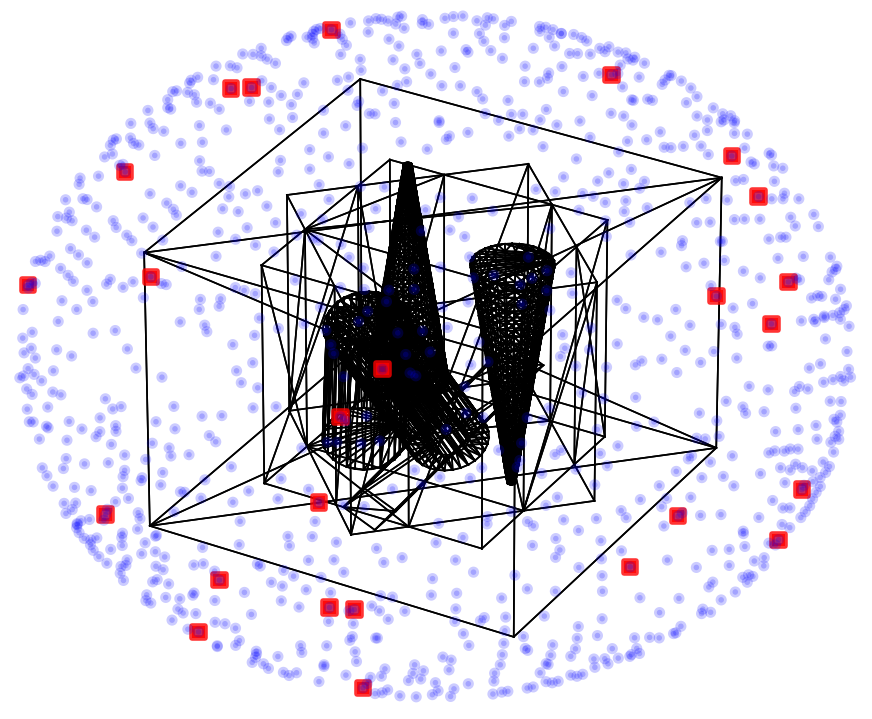}}
    \caption{Optimized Projections for Test Specimen in Section \ref{cube}. Red projections represent the 25 selected ones, while blue indicates others.}
    \label{fig:experiment}
\end{figure} 

A promising approach for performing CT trajectory optimization in terms of data completeness was introduced by Herl et al. \cite{herl2022xray}. Their approach merges the Tuy-Smith condition with the Nyquist-Shannon theorem to create a comprehensive data completeness condition. This condition can evaluate the completeness of any set of projections, making it suitable for complex scanning scenarios. However, since its calculation of data completeness is based on a greedy approach, it is highly dependent on the projections already chosen.


To overcome this limitation, we introduce an integer optimization-based (IP) approach to assess data completeness, focusing on optimizing the coverage of discretely sampled unit spheres around voxels of interest (VOIs). We employ a branch-and-cut optimization technique to improve solution quality. In addition, we can quantify the solution quality by an "optimality gap", which is the gap between the best possible objective and the best found objective. Our experiments highlight the advantages of our automated IP-based projection selection method over traditional approaches. It offers the possibility to optimize heuristically found solutions and provides a deep insight into the complexity of CT trajectory optimization.

\section{Methods}

\subsection{Continuous CT Data Completeness Condition}

Computed Tomography (CT) relies on the Radon transform, a mathematical tool for translating spatial data into the Radon space. 
%
%
%
To ensure complete CT reconstruction, it is essential for the Radon space to exhibit continuity, a concept initially introduced by Radon.

In 1983 \cite{tuy1983inversion}, Tuy introduced data completeness conditions for CT reconstruction based on continuous curves. A curve $C$ is defined as a continuous function $C: \Theta \rightarrow \mathbb{R}^3$, where $\Theta$ represents an interval in $\mathbb{R}$. The unit sphere $\mathbb{S} \subset \mathbb{R}^3$ and $\Omega$, a compact region encompassing the object characterized by the density function $f$, play crucial roles in this context.

Tuy's conditions provide the foundation for achieving data completeness in CT reconstruction. However, these conditions rely on two assumptions that, while theoretically valid, pose practical challenges. The first assumption is the \emph{Detector Resolution Assumption} \cite{herl2022xray}, in which Tuy assumes precise knowledge of the impact position of each measured photon, implying the need for a detector with infinitely small pixels. The second assumption, known as the \emph{Source Assumption} \cite{herl2022xray}, postulates that X-rays are emitted at every point along the source trajectory, effectively resulting in infinite projections.

Three critical conditions must be met to ensure accurate reconstruction of an object within the defined region $\Omega$ using data gathered along curve $C$. Firstly, curve $C$ should avoid entering or intersecting $\Omega$, remaining entirely outside of it. Secondly, curve $C$ should exhibit desirable characteristics such as continuity, boundedness, and smoothness to ensure its predictable behavior. Lastly, at any point within $\Omega$ and in any given direction represented by the unit sphere $\mathbb{S}$, a parameter $\theta$ must exist linked to curve $C$. This parameter aligns with the object's information in the specified direction. Together, these conditions facilitate precise object reconstruction from projection data.

In 1985 \cite{smith1985image}, Smith introduced a simplified condition known as the \emph{Tuy-Smith Condition}, which remains sufficient for CT reconstruction under assumptions 1 and 2 (Detector Resolution and Source Assumption). According to the Tuy-Smith Condition, an object within region $\Omega$ can be accurately reconstructed from projection data generated along curve $C$ if, for all $(x, u) \in \Omega \times \mathbb{S}$, there exists $\theta \in \Theta$ such that $x \cdot u = C(\theta) \cdot u$.

The Tuy-Smith Condition asserts that every plane intersecting the region of interest must also intersect the source trajectory. Given that each point in the Radon space corresponds to a plane in the spatial domain, adhering to this condition guarantees comprehensive sampling of the Radon space.

As both the detector resolution assumption and the source assumption are impractical, the projections, as well as the continuous trajectory, require sampling. The Nyquist-Shannon sampling theorem is applied to ensure adequate information. This theorem dictates that, for faithful CT reconstruction, the sampling rate must be at least twice the highest frequency within the object, corresponding to fine details or sharp transitions in the object's density distribution.

\subsection{Discrete CT Data Completeness Condition}
For archiving a discrete formulation of CT data completeness, we follow the ideas of \cite{maier2015discrete}, which were extended by \cite{herl2022xray}. First, we require a discrete sampling of the unit sphere in vectors $ (u) \in (\mathbb{S}_u) $. Following ideas from \cite{gonzalez2009measurement}, we sample the sphere as uniformly as possible. It's important to note that we only need to sample half of the unit sphere, as the plane integrals exhibit symmetry for the top and bottom portions.

Next, we need to compute the data completeness in terms of Radon sphere coverage for every voxel of interest (VOI). Our objective is to identify vectors from the unit sphere \(u\) that are covered by the normal vectors of the Radon planes, i.e., vectors that are perpendicular to the current viewing direction \(d \in D\), which represents the connection between the source and the current voxel. We only consider those vectors \(d\) that actually hit the detector.

The Nyquist-Shannon sampling theorem states that in Radon space, the angular separation between adjacent planes need only be less than the maximum angular gap  $ \Delta \gamma $ \cite{herl2022xray}. In addition, since the inner product of two unit vectors describes the cosine of the angle between them, and we want to find vectors that deviate by less than \( \Delta \gamma \), the following, already reformulated, condition holds for the vectors of interest \cite{herl2022diss}: 
\begin{equation}
	|d^Tu| < \sin ( \Delta \gamma )
	\label{hit_check}
\end{equation}
It's worth noting that the sine function can be precomputed since \(\Delta \gamma \) remains constant in our case. The above equation is evaluated for every unit vector and every viewing direction through the VOI.

\subsection{Integer Program for CT Trajectory Optimization}
The discrete formulation of a CT data completeness condition is used to model an integer program, which allows us to find an optimized set of $ k $ viewing directions $ D_{k} \subseteq D $. Therefore, we sample a unit sphere in vectors $ u \in \mathbb{S}_u^{V} $ for a VOI $V $. Using Equation \eqref{hit_check}, we precompute for every direction $ d \in D $  which unit vectors $ u \in \mathbb{S}^{V}  $ fulfills Equation \eqref{hit_check}. With that information, we build up a matrix $ S_{V} \in\{0,1\}^{\lvert D \rvert \times \lvert \mathbb{S}^{V} \rvert}$, where an entry of 1 indicates that Equation \eqref{hit_check} holds. Generally, this formulation can be extended to multiple VOIs by sampling a unit sphere for each VOI.

Additionally, we use the idea of \cite{herl2020scanning} to limit the number of selectable viewing directions $ d $ by incorporating an absorption-based metric into our model. Therefore, we select a threshold $\alpha$ indicating the percentage of photons still reaching the detector after passing through the object. The absorption for a viewing direction is evaluated in the projection domain by calculating the average normalized pixel value in the regions of interest indicated by the voxels of interest. This helps us assess the viewing direction's reliability for the respective VOIs. This metric can be precomputed, and the results are saved in a $ \lvert D \rvert \times 1 $-matrix, denoted by $ A_D $. Instead of absorption, alternative projection-based metrics constrained by threshold values can be employed,  increasing the optimization process's flexibility and enabling task-specific adjustments.

Combining all the information acquired, we formulate the selection $x$ of $k$ viewing directions as a convex integer optimization program. Our objective is to maximize the coverage $C = \sum_{i=1}^{\lvert \mathbb{S}^{V} \rvert} c_i $ of the unit sphere. 
\begin{align}
	\text{Maximize:} \quad & \sum_{i=1}^{\lvert \mathbb{S}^{V} \rvert} c_i \\
	\text{Subject to:} \quad & \sum_{i=1}^{\lvert D \rvert} x_i = k \\
	& A_D \cdot x \leq \alpha\\
	& c \leq {S_{V} }^{T}\cdot x \\
	& x,c \in \{0,1\}.
\end{align}


\subsection{Reference Methods}
The effectiveness of the proposed method is compared to two reference methods. First, we use an equidistant circular CT trajectory as a reference since this is one of the state-of-the-art CT trajectories for non-destructive testing. To make a fair comparison, we chose the same number $ k $ of viewing directions as the IP-based approach. Because of the equidistant sampling, the absorption-based metric is not considered. The second reference method selects $ k $ viewing direction with a greedy approach, trying to maximize the coverage of the unit sphere vectors $ \mathbb{S}_u^{V} $ by selecting the viewing direction $ d $, which contributes most to the overall coverage $ C $.

\section{Experiments and Results}
We used Python and the Gurobi optimizer for our CT trajectory optimization. Given the potential exponential runtime of integer programs, we incorporated a callback method into the optimizer. This method monitors the progress of the optimization and intervenes if necessary. Specifically, it checks whether significant progress has been made in reducing the optimality gap within the last \SI{20}{\second}. If the improvement fell below \num{1e-8}, the optimization was terminated prematurely, even if this resulted in a suboptimal solution. This decision is consistent with the observation that extended optimization efforts often yield negligible improvements within a reasonable timeframe, emphasizing timely results while maintaining acceptable solution quality.

Both experiments were evaluated using identical metrics, comparing the reconstructed volume between the sparse CT trajectory and the trajectory including all available projections. For reconstruction we used the Algebraic Reconstruction Technique because of its capability to reconstruct arbitrary orbits. The quantitative evaluation included the calculation of the Structural Similarity Index (SSIM), the contrast-to-noise ratio (CNR) and the Peak Signal to Noise Ratio (PSNR) within the ROI. 

\subsection{CT Trajectory Optimization for Metal Artifact Reduction} \label{expA}
\begin{figure}[!t]
    \centering
    \subfloat[]{\includegraphics[height=1.2in]{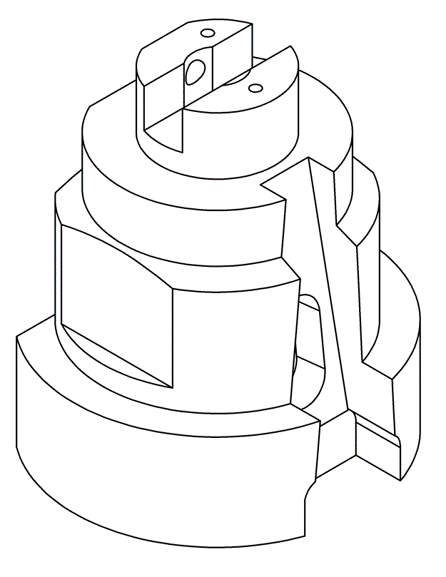}}
    \hfil
    \subfloat[]{\includegraphics[height=1.2in]{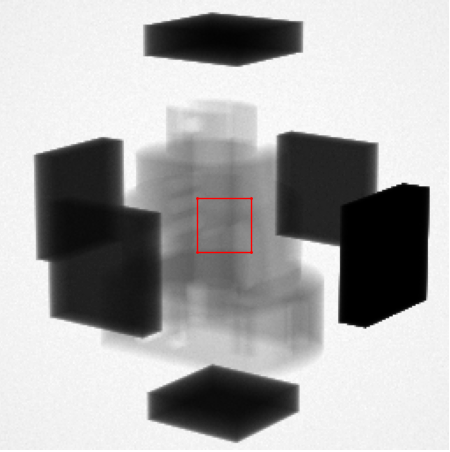}}
    
    \caption{Overview of the test specimen of Experiment \ref{expA}. (a) Sketch of the carbon test specimen. (b) Simulated projection of the test specimen, red box markes the projected VOI \cite{herl2022diss}.}
    \label{fig:experiment_setup_2}
\end{figure}
In this study, we aimed to visualize the internal structure of a \SI{3}{\centi\meter} carbon sample, as shown in Figure~\ref{fig:experiment_setup_2}. To add complexity, we placed the carbon sample between six \SI{1.5}{\centi\meter} wide and \SI{0.4}{\centi\meter} thick iron plates, resulting in a highly attenuating and symmetrical configuration. This setup, similar to a previous experiment in \cite{herl2022diss}, aimed to showcase the effectiveness of a CT trajectory optimization approach. Our dataset consisted of 51 tilt angles ranging from \SI{-90}{\degree} to \SI{90}{\degree} relative to the horizontal axis, enabling projections over a half-sphere. For each tilt angle, we simulated 61 evenly spaced projections within a \SI{216}{\degree} range. The simulations incorporated a polychromatic spectrum, simulated noise, and excluded scattering effects. Scan parameters included a \SI{150}{\kilo\volt} voltage, no pre-filter, a pixel size of \SI{700}{\micro\meter}, a \SI{18}{\centi\meter} $\times$ \SI{18}{\centi\meter} detector, and a magnification factor of 3.

After examining the simulated projections, we adjusted the $\alpha$ parameter to exclude X-rays with excessive attenuation, particularly those passing through metal plates in our experimental setup. This modification set $\alpha$ to a value of 0.7, indicating that, on average, at least 30\% of the photons in our region of interest (ROI) successfully penetrated the object. Consequently, this constraint reduced the available selectable projections from 3111 to 1851.
For our projection selection method, we aimed to select a set of 61 projections, corresponding to the number of viewpoints in each \SI{216}{\degree} circle. To facilitate comparative analysis, we chose the untilted circle trajectory as the most fundamental within our projection set. We chose the VOI as the center of the object and sampled a unit sphere around it, resulting in $\lvert \mathbb{S}_u^{V} \rvert = 2000 $ unit sphere points and a maximum angular distance of $\Delta \gamma = 0.01$, a choice informed by the methodology outlined in \cite{herl2022xray}.
\begin{table}[bhtp]
\centering
\caption{Comparison of Quality Metrics for 61 Projections}
\begin{tabular}{@{}ccccc@{}}
\toprule
\textbf{Approach} & \textbf{SSIM}$\uparrow$ & \textbf{PSNR}$\uparrow$ & \textbf{CNR}$\uparrow$ &\textbf{Coverage}$\uparrow$ \\ 
\midrule
Circular & 0.60 & 118.7db  & 1.12 & 45\% \\
Greedy &  0.62 & 118.8db & 3.3& 64\%  \\ 
IP & 0.72 &  119.9db & 4.4& 91\%  \\
\bottomrule
\end{tabular}%
\label{tab:metricsA}
\end{table} 

The IP approach performs best in terms of image quality and data completeness and has a higher SSIM (0.72) and PSNR (119.9 dB) than the other methods. In addition, it is better at selecting projections, covering 91\% of the unit sphere, while Greedy achieves only 64\%. The small optimality gap of 1.79\%  underlines the IP approach's potential as a promising choice for CT trajectory design optimization.


\subsection{CT Trajectory Optimization for an Arbitrary VOI}\label{cube}

We artificially generated an empty cube for our second experiment and placed multiple geometric shapes inside, as visible in Figure \ref{fig:experiment}. We simulated 1000 projections of a full sphere with the object placed in the center to ensure variable viewing positions. Our simulations incorporated a polychromatic spectrum and simulated noise, omitting the simulation of scattering effects. Additional scan parameters included a voltage of \SI{150}{\kilo\volt}, no pre-filter, a pixel size of \SI{600}{\micro\meter}, a detector size of \SI{15.4}{\centi\meter}$\times$\SI{15.4}{\centi\meter}, and a magnification factor of 2. The imaging task was to enhance reconstruction quality for the cylindrical object inside the cube. After simulating the projections, we selected $\alpha = 0.84$ to exclude high attenuation X-rays. This resulted in a reduction of the available projections from 1000 to 535. 

We evaluated our method on $k = 25$ projections. We simulated an equidistantly sampled \SI{360}{\degree} circular trajectory for comparison. To test the performance of the method on non-central VOIs, we chose a voxel on the cylinder shaft as the VOI. We sampled a unit sphere around the VOI with $\lvert \mathbb{S}_u^{V} \rvert = 1500 $ points and  a maximum angular distance of $\Delta \gamma = 0.015$.

\begin{figure}
		
    \begin{tabular}{r m{1.5cm} m{1.5cm} m{1.5cm}  m{1.5cm}}
\multicolumn{1}{c}{} & \multicolumn{1}{c}{\centering Reference} & \multicolumn{1}{c}{\centering Circular} & \multicolumn{1}{c}{\centering Greedy} & \multicolumn{1}{c}{\centering IP} \\ 

\multicolumn{1}{c}{xy} & \adjustbox{valign=m}{\includegraphics[width=1.8cm]{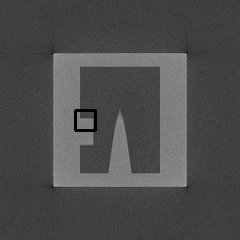}} & \adjustbox{valign=m}{\includegraphics[width=1.8cm]{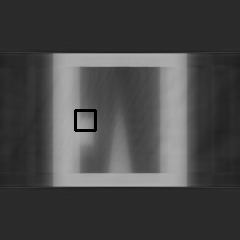}} & \adjustbox{valign=m}{\includegraphics[width=1.8cm]{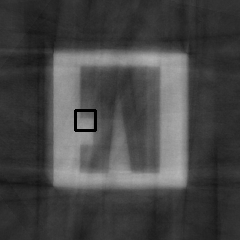}} & \adjustbox{valign=m}{\includegraphics[width=1.8cm]{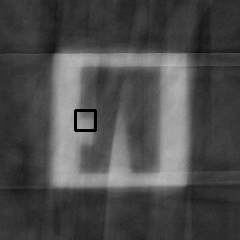}} \\ 

\multicolumn{1}{c}{xz} & \adjustbox{valign=m}{\includegraphics[width=1.8cm]{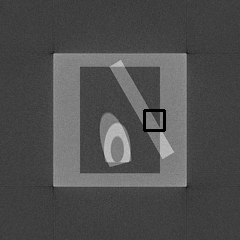}} & \adjustbox{valign=m}{\includegraphics[width=1.8cm]{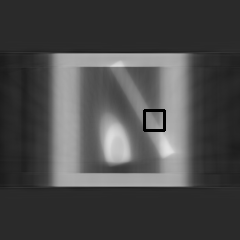}} & \adjustbox{valign=m}{\includegraphics[width=1.8cm]{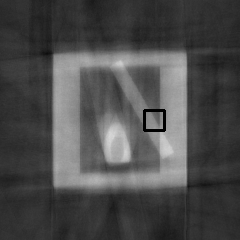}} & \adjustbox{valign=m}{\includegraphics[width=1.8cm]{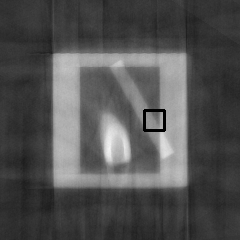}} \\ 

\multicolumn{1}{c}{yz} & \adjustbox{valign=m}{\includegraphics[width=1.8cm]{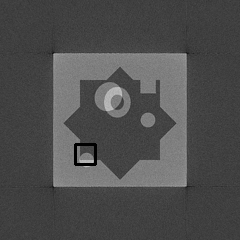}} & \adjustbox{valign=m}{\includegraphics[width=1.8cm]{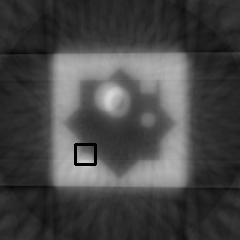}} & \adjustbox{valign=m}{\includegraphics[width=1.8cm]{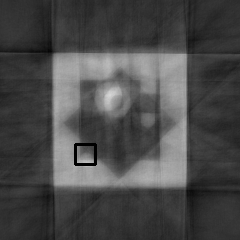}} & \adjustbox{valign=m}{\includegraphics[width=1.8cm]{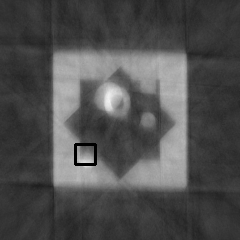}} \\ 

\end{tabular}
  
	\caption{Optimized CT trajectory for the marked region. The optimized and circular CT trajectories consist of $k =25$ projections, while the reference data was a spherical CT trajectory with 1000 projections.}
	\label{fig:complexCubeWorstCase}
\end{figure}

\begin{table}[htbp]
\centering
\caption{Comparison of Quality Metrics for 25 Projections}
\begin{tabular}{@{}ccccc@{}}
\toprule
\textbf{Approach} & \textbf{SSIM}$\uparrow$ & \textbf{PSNR}$\uparrow$ & \textbf{CNR}$\uparrow$ &\textbf{Coverage}$\uparrow$ \\ 
\midrule
Circular & 0.5 & 113.7db  & 19 & 34.5\% \\
Greedy &  0.57 & 115.3db & 26& 48.1\%  \\ 
IP & 0.59 &  113.9db & 48& 73.6\%  \\
\bottomrule
\end{tabular}%
\label{tab:metrics_b}
\end{table}
The results presented in table \ref{tab:metrics_b} and figure \ref{fig:complexCubeWorstCase} highlight the potential of the IP approach as a strong contender. It achieves a respectable SSIM of 0.59 and a decent PSNR of 113.9dB, indicating competitive image quality compared to the circular and greedy methods. Furthermore, in terms of CNR, IP outperformed its counterparts, demonstrating its ability to improve image contrast. It also excelled in coverage, covering 73.6\% compared to 48.1\% of the unit sphere. With a narrow optimality gap of 0.75\%, the IP approach demonstrates its ability to provide near-optimal solutions. 

\section{Conclusion}
In this study, we introduced an IP-based CT trajectory optimization method to address the challenges of data completeness and projection selection in CT. Our approach uses discrete data completeness conditions to formulate an optimization problem for selecting an optimized set of viewing directions. This method ensures data completeness while considering absorption-based metrics for reliable evaluation of the selected projections.

Experimental results demonstrate the effectiveness of our approach in improving the quality of CT reconstructions. In both experiments, our optimized CT trajectory outperformed the standard and greedy trajectory regarding image quality. In addition, we used the concept of an optimality gap to quantify optimality in terms of our discrete data completeness formulation. The experiments emphasized the importance of careful projection selection.

Future work includes studying parameter influence, refining the IP formulation for faster convergence, and exploring additional projection-based metrics for CT trajectory optimization.

\section*{Acknowledgment}
This research was financed by the ``SmartCT – Methoden der Künstlichen Intelligenz für ein autonomes Roboter-CT System'' \ project (project nr. DIK-2004-0009).

\ifCLASSOPTIONcaptionsoff
  \newpage
\fi

\end{document}